\documentclass[journal,twoside,web]{ieeecolor}
\usepackage{generic}
\usepackage{amsmath,amssymb,amsfonts}
\usepackage{algorithmic}
\usepackage{graphicx}
\usepackage{textcomp}

\usepackage{multirow}
\usepackage{booktabs}
\usepackage{siunitx}
\usepackage{float}
\usepackage{array}
\usepackage{subcaption}
\usepackage{bm}
\usepackage{tabularx} 
\usepackage[hidelinks]{hyperref}
\usepackage[
    backend=biber,
    style=ieee,
    sorting=none,
    doi=false,
    url=false,
    isbn=false,
    maxnames=6, 
    minnames=1
]{biblatex}
\def\BibTeX{{\rm B\kern-.05em{\sc i\kern-.025em b}\kern-.08em
    T\kern-.1667em\lower.7ex\hbox{E}\kern-.125emX}}
\usepackage{fancyhdr}
\emergencystretch=\maxdimen
\begin{document}
\title{\textbf{MAC-Net: A Multi-Task Deep Learning Framework for Modeling Cognitive Function From Task-Based fMRI}}
\author{
Md. Tanvir Rahman,
Nabil Anan Orka,
Asaduzzaman Khan, and
Mohammad Ali Moni
\thanks{Md. Tanvir Rahman, Nabil Anan Orka, Asaduzzaman Khan, and Mohammad Ali Moni are with the School of Health and Rehabilitation Sciences, The University of Queensland, QLD 4072, Australia.}
\thanks{Md. Tanvir Rahman is also with the Department of Information and Communication Technology, Mawlana Bhashani Science and Technology University, Tangail 1902, Bangladesh.}
}

\maketitle
\thispagestyle{fancy} 

\begin{abstract}
Objective assessment of cognitive function from neural signals supports neurorehabilitation and human-performance monitoring, but individual-level prediction from task-based functional magnetic resonance imaging (tfMRI) remains difficult because neural features coexist with substantial demographic and scanner-related variation. This study presents a covariate-aware multi-task deep learning framework, the Multi-task Activation and Contrast Network (MAC-Net), for modeling individual cognitive function from regional tfMRI representations. By structurally isolating task-fMRI features into a dedicated neural pathway and restricting participant variables to a terminal late-fusion pathway, MAC-Net provides a generalizable blueprint to prevent dominant covariates from suppressing high-dimensional clinical representations during feature learning. Baseline data from 6,500 Adolescent Brain Cognitive Development Study participants were evaluated using fixed family-aware cross-validation. MAC-Net was benchmarked against regularized linear models, random forests, and alternative deep-learning architectures. The N-back plus Monetary Incentive Delay configuration achieved coefficients of determination (\(R^2\)) of 0.174, 0.238, and 0.277 for fluid, crystallized, and total cognitive scores, respectively. MAC-Net outperformed the strongest covariate-only baseline (0.178) and alternative deep models (0.217) for total cognition. N-back was the most informative paradigm, whereas incorporating the Stop Signal Task marginally degraded performance, likely due to the curse of dimensionality when integrating non-informative features. Feature attributions via Integrated Gradients, DeepLIFT, and Input Gradient were highly concordant, localizing working-memory-related frontal, parietal, and cingulate regions. These findings demonstrate that covariate-aware multi-task modeling yields reproducible, individualized cognitive-function estimations, establishing a robust neural engineering framework for clinical translation.
\end{abstract}


\begin{table*}[t]
\caption{Participant characteristics and cognitive-score distributions across the training, validation, and test partitions of the five family-aware cross-validation folds.}
\label{tab:demographics}
\centering
\small
\setlength{\tabcolsep}{4pt}
\begin{tabular*}{\textwidth}{@{\extracolsep{\fill}} l l l l l l c c c}
\toprule
\textbf{Fold} & \textbf{Split} & \textbf{N} & \textbf{Males} & \textbf{Females} & \textbf{Age (months)} & \textbf{\(G_f\)} & \textbf{\(G_c\)} & \textbf{\(G_t\)} \\
\midrule
\multirow{3}{*}{1} & Train & 3891 & 1947 & 1944 & 119.63 \(\pm\) 7.53 & 94.05 \(\pm\) 9.44 & 87.82 \(\pm\) 6.50 & 88.57 \(\pm\) 7.97 \\
  & Val   & 1303 & 682 & 621 & 119.48 \(\pm\) 7.54 & 93.84 \(\pm\) 9.70 & 87.72 \(\pm\) 6.51 & 88.39 \(\pm\) 8.20 \\
  & Test  & 1306 & 638 & 668 & 119.84 \(\pm\) 7.59 & 93.64 \(\pm\) 9.56 & 87.84 \(\pm\) 6.47 & 88.36 \(\pm\) 8.03 \\
\midrule
\multirow{3}{*}{2} & Train & 3891 & 1930 & 1961 & 119.78 \(\pm\) 7.53 & 93.99 \(\pm\) 9.43 & 87.87 \(\pm\) 6.49 & 88.58 \(\pm\) 7.99 \\
  & Val   & 1306 & 655 & 651 & 119.39 \(\pm\) 7.59 & 93.82 \(\pm\) 9.57 & 87.69 \(\pm\) 6.47 & 88.33 \(\pm\) 7.99 \\
  & Test  & 1303 & 682 & 621 & 119.48 \(\pm\) 7.54 & 93.84 \(\pm\) 9.70 & 87.72 \(\pm\) 6.51 & 88.39 \(\pm\) 8.20 \\
\midrule
\multirow{3}{*}{3} & Train & 3897 & 1971 & 1926 & 119.77 \(\pm\) 7.50 & 93.87 \(\pm\) 9.54 & 87.81 \(\pm\) 6.47 & 88.47 \(\pm\) 8.05 \\
  & Val   & 1297 & 641 & 656 & 119.53 \(\pm\) 7.63 & 94.21 \(\pm\) 9.36 & 87.89 \(\pm\) 6.58 & 88.72 \(\pm\) 8.00 \\
  & Test  & 1306 & 655 & 651 & 119.39 \(\pm\) 7.59 & 93.82 \(\pm\) 9.57 & 87.69 \(\pm\) 6.47 & 88.33 \(\pm\) 7.99 \\
\midrule
\multirow{3}{*}{4} & Train & 3915 & 1975 & 1940 & 119.57 \(\pm\) 7.57 & 93.77 \(\pm\) 9.61 & 87.75 \(\pm\) 6.49 & 88.36 \(\pm\) 8.07 \\
  & Val   & 1288 & 651 & 637 & 119.98 \(\pm\) 7.37 & 94.14 \(\pm\) 9.37 & 87.88 \(\pm\) 6.43 & 88.66 \(\pm\) 7.92 \\
  & Test  & 1297 & 641 & 656 & 119.53 \(\pm\) 7.63 & 94.21 \(\pm\) 9.36 & 87.89 \(\pm\) 6.58 & 88.72 \(\pm\) 8.00 \\
\midrule
\multirow{3}{*}{5} & Train & 3906 & 1978 & 1928 & 119.47 \(\pm\) 7.59 & 93.95 \(\pm\) 9.54 & 87.76 \(\pm\) 6.52 & 88.48 \(\pm\) 8.06 \\
  & Val   & 1306 & 638 & 668 & 119.84 \(\pm\) 7.59 & 93.64 \(\pm\) 9.56 & 87.84 \(\pm\) 6.47 & 88.36 \(\pm\) 8.03 \\
  & Test  & 1288 & 651 & 637 & 119.98 \(\pm\) 7.37 & 94.14 \(\pm\) 9.37 & 87.88 \(\pm\) 6.43 & 88.66 \(\pm\) 7.92 \\
\bottomrule
\end{tabular*}
\end{table*}

\section{Introduction}
Cognitive function is a fundamental component of human performance that influences learning, independence, and recovery across a wide range of neurological conditions \cite{belkacem2023closed}. Objective characterization of cognitive capacity is therefore a critical target for neural engineering, particularly because behavioral tests often provide an incomplete representation of an individual's underlying neural state \cite{young2024covert}. Recent works have demonstrated the potential of computational analysis of task-related neural signals for cognitive evaluation and rehabilitation-oriented prediction, including task-state EEG-based cognitive assessment and machine-learning approaches for MRI-based dementia prediction and treatment-response estimation \cite{wen2022taskstate,kuo2023optimized,lee2024machine,gonzalezlima2025machine}. These studies highlight the broader potential of neural-signal-based systems to provide objective measures that complement conventional functional assessment.

Crucially for neural engineering, standard behavioral evaluations rely heavily on intact motor skills and verbal communication. These expressive pathways are frequently compromised in severe neurological conditions, such as stroke or traumatic brain injury, which can mask a patient's true cognitive preservation \cite{schnakers2020covert, young2024covert}. By bypassing these physical bottlenecks, computational neural-signal estimators offer a direct, objective read-out of cognitive capability \cite{schnakers2020covert}. Furthermore, individualized neural state estimation provides a continuous, quantitative biomarker of neuroplasticity \cite{jin2024electroencephalogram}. This capability is essential for developing closed-loop rehabilitation technologies, such as adaptive neurofeedback or targeted non-invasive brain stimulation, that dynamically adjust to a patient's actual neural recovery trajectory over time \cite{belkacem2023closed, jin2024electroencephalogram}.

Task-based functional magnetic resonance imaging (tfMRI) provides a noninvasive means of characterizing neural responses elicited by controlled cognitive states. Unlike resting-state measurements, task paradigms explicitly engage functional systems associated with working memory, attention, reward processing, and response inhibition, thereby providing experimentally defined neural contexts in which individual differences can be examined. Previous neuroimaging studies have demonstrated that task-evoked brain states improve the prediction of individual traits \cite{greene2018task,gal2022traits}, and that integrating tfMRI information across multiple paradigms boosts predictive reliability relative to relying on a single task state \cite{tetereva2022capturing}. These findings motivate computational frameworks that can integrate heterogeneous task-evoked representations while preserving information relevant to individual cognitive variation.

A major challenge in developing such systems, particularly in large-scale, multisite developmental cohorts such as the Adolescent Brain Cognitive Development (ABCD) Study, is that tfMRI measurements are heavily influenced by non-neural factors \cite{abcd,abcd_guideline}. Biological sex, intracranial volume, scanner hardware, and head motion introduce structured dependencies that can artificially inflate predictive performance \cite{marek2022reproducible,gell2024measurement}. This issue is critical for translational neural engineering, where models must demonstrate predictive value beyond readily available demographic variables. Consequently, preventing information leakage requires rigorous control of confounding variables, family-aware partitioning, and fold-specific preprocessing \cite{scheinost2019rules,snoek2019confounds}. Furthermore, strong predictive performance alone does not guarantee a unique biological mechanism, as varying feature-selection methods can yield divergent neurobiological interpretations \cite{adkinson2026feature}.

A second challenge concerns the representation and integration of high-dimensional tfMRI data. Structured feature spaces spanning multiple cognitive states require flexible nonlinear function approximation. While deep learning offers a natural mechanism for this integration, increasing architectural complexity does not necessarily yield improved prediction. Regularized linear models remain highly competitive in neuroimaging \cite{schulz2020different,pat2023explainable}, and recent benchmarks demonstrate that simpler architectures often outperform complex alternatives under realistic evaluation conditions \cite{popov2024simple,han2026rethinking}. These observations necessitate direct architectural benchmarking rather than assuming greater model complexity inherently produces better neural-signal representations. Indeed, systematic evaluations of deep learning in neurocognition consistently identify inconsistent evaluation practices and the limited use of explainability methods as primary hurdles to clinical translation \cite{rahman2025review}.

A third challenge is interpreting model-derived neural features. In neural engineering applications, predictive accuracy alone is insufficient when the objective is to understand which aspects of a neural signal contribute to an estimated functional outcome. Explainability methods provide post hoc information regarding model sensitivity, but their results depend heavily on the attribution algorithm, reference baseline, and feature correlations \cite{kohoutova2020interpreting}. Consequently, reliable interpretation requires assessing attribution consistency across complementary methods and independent data partitions rather than relying on a single saliency map.

To address these interconnected challenges, we developed the Multi-task Activation and Contrast Network (MAC-Net), a covariate-aware deep-learning framework for estimating individual cognitive function from regional tfMRI activations. MAC-Net analyzes tfMRI data and participant variables separately, combining demographic and acquisition covariates only at the final fusion stage. Using data from 6,500 ABCD Study participants, we quantified cognitive abilities via NIH Toolbox measures across three task paradigms: working memory (N-back), reward processing (Monetary Incentive Delay, MID), and response inhibition (Stop Signal Task, SST). This study addresses three core engineering questions. First, does tfMRI provide predictive information beyond prespecified covariates when evaluated under leakage-resistant, family-aware cross-validation? Second, does integrating complementary task paradigms improve estimation relative to individual tasks? Third, does MAC-Net remain competitive against conventional and alternative deep learning architectures while yielding stable, multi-method feature attributions via Integrated Gradients, DeepLIFT, and Input\(\times\)Gradient?

The objective of this work is not to present a clinical diagnostic system, but to establish a rigorous neural-signal modeling framework for cognitive-function estimation. By combining multi-task fMRI representations, explicit covariate-aware fusion, leakage-resistant evaluation, architectural benchmarking, and attribution consensus, this study provides a methodological foundation for future investigation in neurological and neurorehabilitation populations, where objective and individualized characterization of cognitive function may complement established behavioral and clinical assessments.

\begin{figure*}[t]
    \centering
    \includegraphics[width=\textwidth]{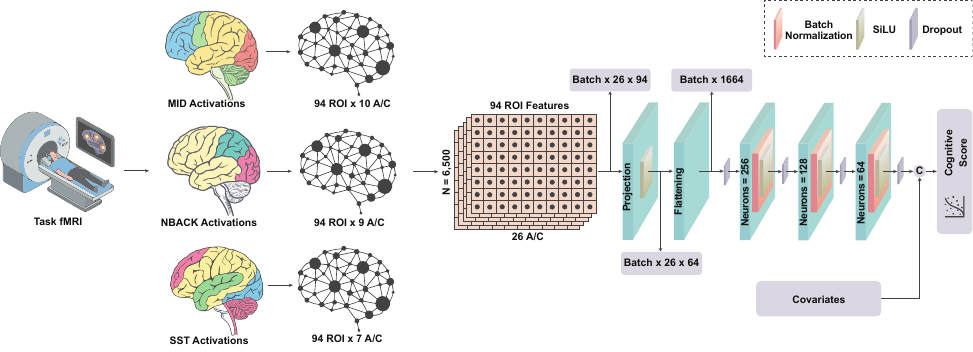}
    \caption{Schematic of the MAC-Net architecture. Regional tfMRI beta estimates (\(C\) conditions \(\times\) 94 regions) are projected into a 64-dimensional embedding, flattened, and processed through a feed-forward backbone (256 \(\rightarrow\) 128 \(\rightarrow\) 64 units) with batch normalization, SiLU, and dropout. Participant and acquisition covariates bypass the tfMRI extractor via terminal late fusion. The primary N-back+MID configuration utilizes \(C=19\) inputs; the N-back+MID+SST sensitivity ablation utilizes \(C=26\).}
    \label{fig:macnet_architecture}
\end{figure*}

\section{Methods}
\label{sec:methods}
\subsection{Cohort and Data Acquisition}
Data were obtained from the Adolescent Brain Cognitive Development (ABCD) Study Tabulated Data Release v5.1 \cite{abcd}. Participants were required to pass strict imaging quality control, possess the ABCD-recommended inclusion flags for the evaluated task paradigms, and pass the FreeSurfer structural protocol \cite{abcd_guideline}. The final complete-case cohort comprised \(N = 6{,}500\) adolescents with complete cognitive outcomes, covariates, and regional tfMRI features. 

To prevent data leakage due to shared genetic or environmental variance, we employed a strict, family-aware, five-fold cross-validation framework \cite{scheinost2019rules}. The five fixed, non-overlapping partitions ensured that siblings and twins were grouped exclusively within the same fold. During each iteration, three groups were used for training, one for validation and early stopping, and one as the held-out test set. Detailed participant characteristics and cognitive-score distributions across these data partitions are provided in Table \ref{tab:demographics}

\subsection{Cognitive Phenotyping and Covariate Integration}
\begin{figure*}[t]
    \centering
    \includegraphics[width=\textwidth]{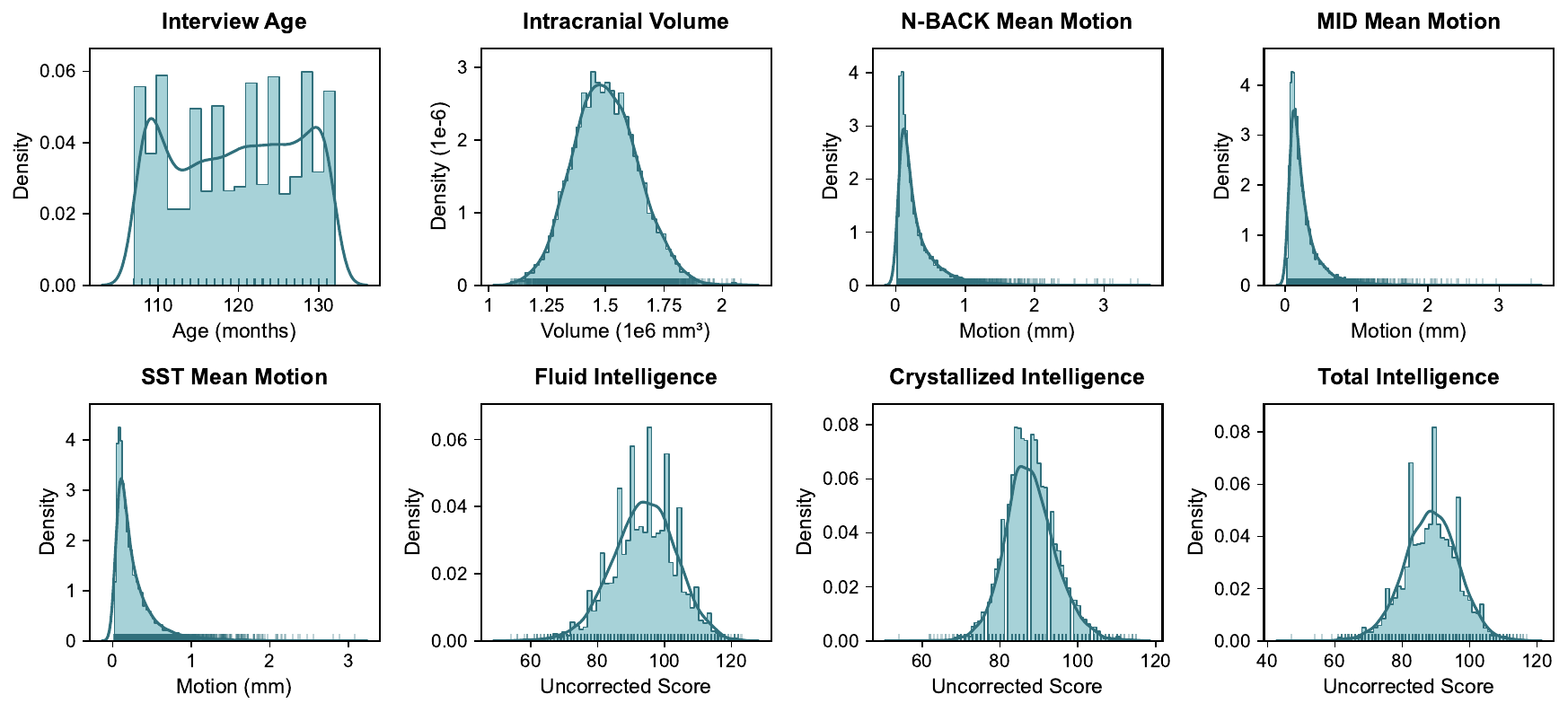}
    \caption{Distributions of continuous covariates and uncorrected NIH Toolbox cognitive outcomes. Histograms and kernel density estimates denote interview age, estimated total intracranial volume (eTIV), task-specific mean framewise displacement, and fluid, crystallized, and total intelligence scores. Categorical covariates (sex, MRI device) are omitted from visualization.}
    \label{fig:distribution}
\end{figure*}

We modeled three continuous targets from the NIH Toolbox Cognition Battery: fluid intelligence (\(G_f\)), crystallized intelligence (\(G_c\)), and total composite intelligence (\(G_t\)) \cite{abcd}. Standard uncorrected scores were utilized to preserve developmental variability, and optimization was stabilized by standardizing (\(Z\)-scoring) the targets exclusively using parameters derived from the training distribution within each fold. The underlying population distributions of these continuous cognitive outcomes and the measured participant covariates are visualized in Fig. \ref{fig:distribution}.

To isolate incremental neural signals from known confounders, the model explicitly incorporated a covariate vector \(\mathbf{c}_i\) containing: interview age (months), biological sex, MRI device serial number, estimated total intracranial volume (eTIV), and task-matched mean framewise displacement \cite{scheinost2019rules,snoek2019confounds}. Continuous covariates were standardized, and categorical variables were one-hot encoded strictly using the training partition to prevent leakage. 

\subsection{Task-fMRI Feature Engineering}
Task activation (beta) estimates and functional contrasts were obtained from the standardized ABCD processing pipeline \cite{abcd_guideline}, averaged across 94 anatomical regions (68 Desikan--Killiany cortical parcels and 26 bilateral ASEG subcortical segments). A complete enumeration of these regions is provided in Supplementary Section S1. 

We extracted a comprehensive suite of main effects and functional contrasts representing distinct psychological processes across three fMRI task paradigms: N-back, MID, and SST. The resulting neural feature space comprised 9 N-back conditions (846 features), 10 MID conditions (940 features), and 7 SST conditions (658 features), totaling 26 unique task-evoked brain states. A detailed breakdown of these conditions and functional contrasts is provided in Supplementary Table S1.

\subsection{MAC-Net Architecture and Optimization}
We introduce MAC-Net, a deep feed-forward architecture engineered for highly granular tfMRI representations. MAC-Net structurally routes high-dimensional tfMRI inputs and measured covariates through separate pathways, ensuring demographics do not artificially dominate early neural feature extraction. An overview of the proposed MAC-Net processing pathways and late-fusion strategy is illustrated in Fig. \ref{fig:macnet_architecture}.

\subsubsection{Deep Feature Extraction}
The flattened tfMRI regional activations first pass through a dedicated projection layer mapping the 94 anatomical regions into a 64-dimensional latent embedding space using Sigmoid Linear Unit nonlinearities. This prevents the model from treating distinct spatial regions and functional tasks uniformly. The projected embeddings subsequently pass through a \(D=3\)-layer feed-forward backbone with progressively halved dimensionality (256 \(\rightarrow\) 128 \(\rightarrow\) 64 neurons). To stabilize gradient flow across the connectomic data, 1D Batch Normalization and aggressive Dropout (\(p = 0.7\)) are applied at every hidden layer.

\subsubsection{Late Fusion of Covariates}
To prevent demographic variables from dominating early feature extraction, MAC-Net employs a late-fusion mechanism. The encoded covariate vector (\(\mathbf{c}_i\)) bypasses the neural extractors and is concatenated exclusively at the terminal readout:
\begin{equation}
    \label{eq:late_fusion}
    \hat{y}_i = \mathbf{W}_{\text{head}} (\mathbf{h}_{\text{final}} \oplus \mathbf{c}_i) + b_{\text{head}}
\end{equation}
where \(\mathbf{h}_{\text{final}}\) represents the deeply processed neural embedding and \(\oplus\) denotes concatenation. This structural constraint forces upstream hidden layers to optimize strictly for functional neural variance.

\subsubsection{Training Protocol}
Model weights were optimized utilizing the AdamW algorithm (maximum learning rate = \(8.7 \times 10^{-4}\), weight decay = \(5 \times 10^{-3}\)) to minimize the Mean Squared Error (MSE) over a maximum of 300 epochs (batch size = 128). A ReduceLROnPlateau scheduler dynamically halved the learning rate upon a 10-epoch validation plateau, and early stopping was enforced with a patience of 30 epochs. Model outputs were inverse-transformed to the original uncorrected score scale before final evaluation to accurately report Mean Absolute Error, MSE, \(R^2\), and Pearson correlation (\(r\)).

\subsection{Ablation Framework and Baselines}
To establish robust benchmarking, MAC-Net was evaluated against standard regressors (Ridge, Lasso, ElasticNet, Random Forest) utilizing the full multimodal feature set and identical family-aware partitions. We also conducted a task-screening ablation (evaluating N-back, MID, and SST individually) and a ``Brain-Only'' ablation (removing \(\mathbf{c}_i\)) to force the architecture to model intelligence solely from neural signals.

To assess MAC-Net's feed-forward integration, we benchmarked it against three state-of-the-art architectures \cite{popov2024simple}: (1) \emph{SpatialGATNet}, utilizing a spatial graph prior (\(K=10\) nearest MNI neighbors) and masked attention; (2) a \emph{Tabular Transformer}, treating anatomical regions as sequence tokens processed via multi-head self-attention; and (3) a \emph{Gated Tabular Network}, employing dynamic feature gating via soft attention and Gated Linear Units.

To guarantee equitable comparison, all advanced baselines were subjected to identical training parameters, family-aware partitions, Dropout (\(p=0.7\)), and the precise late-fusion covariate mechanism (Eq.~\ref{eq:late_fusion}).

\subsection{Multi-Method Attribution Consensus}
Feature attributions were computed for held-out test participants using Integrated Gradients (50 approximation steps), DeepLIFT, and Input\(\times\)Gradient \cite{sundararajan2017ig,shrikumar2017deeplift}. Because features were standardized, a value of zero inherently represented the training-partition mean. Categorical dummy-variable attributions were dynamically aggregated across their respective columns to represent global magnitude.

Absolute attribution values were averaged across participants and min-max normalized to \([0,1]\). A final consensus score was calculated by averaging the normalized scores across the three methods, and methodological convergence was quantified using pairwise Spearman rank correlations. Because covariates entered through late-fusion, the resulting neural maps reflect the fitted model's attributions strictly to its tfMRI inputs.

\begin{table*}[t]
\caption{Baseline modeling performance. The top panel for each target displays covariate-only machine learning baselines (using age, sex, eTIV, scanner, and motion). The bottom panel displays MAC-Net performance using isolated single-task tfMRI inputs along with these common covariates. Values reflect mean \(\pm\) standard deviation across five held-out test folds.}
\label{tab:combined_baselines}
\centering
\small
\setlength{\tabcolsep}{4pt}
\begin{tabular*}{\textwidth}{@{\extracolsep{\fill}} l l c c c c}
\toprule
\textbf{Target} & \textbf{Model / Modality Configuration} & \textbf{\(R^2\) (\(\uparrow\))} & \textbf{Pearson \(r\) (\(\uparrow\))} & \textbf{MAE (\(\downarrow\))} & \textbf{MSE (\(\downarrow\))} \\
\midrule
\multirow{9}{*}{\textbf{\(G_f\)}} 
& \textit{Covariate-Only Baselines} & & & & \\
& \quad \; -- Ridge & 0.113 \(\pm\) 0.006 & 0.338 \(\pm\) 0.008 & 7.091 \(\pm\) 0.073 & 80.257 \(\pm\) 1.652 \\
& \quad \; -- Lasso & \textbf{0.114 \(\pm\) 0.006} & \textbf{0.340 \(\pm\) 0.008} & \textbf{7.083 \(\pm\) 0.075} & \textbf{80.112 \(\pm\) 1.686} \\
& \quad \; -- ElasticNet & 0.113 \(\pm\) 0.005 & 0.338 \(\pm\) 0.008 & 7.089 \(\pm\) 0.091 & 80.207 \(\pm\) 1.867 \\
& \quad \; -- Random Forest & 0.087 \(\pm\) 0.013 & 0.300 \(\pm\) 0.018 & 7.207 \(\pm\) 0.135 & 82.637 \(\pm\) 2.738 \\
& \textit{Single-Task MAC-Net with Covariates} & & & & \\
& \quad \; -- N-back & \textbf{0.178 \(\pm\) 0.014} & \textbf{0.424 \(\pm\) 0.015} & \textbf{6.832 \(\pm\) 0.042} & \textbf{74.349 \(\pm\) 1.585} \\
& \quad \; -- MID & 0.118 \(\pm\) 0.010 & 0.346 \(\pm\) 0.014 & 7.091 \(\pm\) 0.079 & 79.815 \(\pm\) 1.924 \\
& \quad \; -- SST & 0.106 \(\pm\) 0.015 & 0.332 \(\pm\) 0.018 & 7.117 \(\pm\) 0.061 & 80.863 \(\pm\) 1.198 \\
\midrule
\multirow{9}{*}{\textbf{\(G_c\)}} 
& \textit{Covariate-Only Baselines} & & & & \\
& \quad \; -- Ridge & \textbf{0.158 \(\pm\) 0.020} & \textbf{0.398 \(\pm\) 0.025} & \textbf{4.641 \(\pm\) 0.056} & \textbf{35.499 \(\pm\) 1.177} \\
& \quad \; -- Lasso & \textbf{0.158 \(\pm\) 0.020} & \textbf{0.398 \(\pm\) 0.025} & 4.641 \(\pm\) 0.058 & 35.499 \(\pm\) 1.191 \\
& \quad \; -- ElasticNet & 0.154 \(\pm\) 0.019 & 0.394 \(\pm\) 0.027 & 4.650 \(\pm\) 0.058 & 35.676 \(\pm\) 1.172 \\
& \quad \; -- Random Forest & 0.126 \(\pm\) 0.017 & 0.357 \(\pm\) 0.021 & 4.731 \(\pm\) 0.027 & 36.866 \(\pm\) 0.889 \\
& \textit{Single-Task MAC-Net with Covariates} & & & & \\
& \quad \; -- N-back & \textbf{0.223 \(\pm\) 0.022} & \textbf{0.473 \(\pm\) 0.023} & \textbf{4.459 \(\pm\) 0.077} & \textbf{32.767 \(\pm\) 1.353} \\
& \quad \; -- MID & 0.192 \(\pm\) 0.021 & 0.440 \(\pm\) 0.022 & 4.543 \(\pm\) 0.071 & 34.059 \(\pm\) 1.194 \\
& \quad \; -- SST & 0.154 \(\pm\) 0.016 & 0.393 \(\pm\) 0.019 & 4.663 \(\pm\) 0.026 & 35.693 \(\pm\) 1.008 \\
\midrule
\multirow{9}{*}{\textbf{\(G_t\)}} 
& \textit{Covariate-Only Baselines} & & & & \\
& \quad \; -- Ridge & 0.177 \(\pm\) 0.006 & 0.421 \(\pm\) 0.007 & 5.765 \(\pm\) 0.074 & 53.077 \(\pm\) 1.507 \\
& \quad \; -- Lasso & \textbf{0.178 \(\pm\) 0.006} & \textbf{0.423 \(\pm\) 0.007} & \textbf{5.762 \(\pm\) 0.075} & \textbf{52.982 \(\pm\) 1.562} \\
& \quad \; -- ElasticNet & 0.175 \(\pm\) 0.006 & 0.419 \(\pm\) 0.007 & 5.771 \(\pm\) 0.077 & 53.184 \(\pm\) 1.571 \\
& \quad \; -- Random Forest & 0.145 \(\pm\) 0.015 & 0.383 \(\pm\) 0.019 & 5.873 \(\pm\) 0.081 & 55.139 \(\pm\) 1.686 \\
& \textit{Single-Task MAC-Net with Covariates} & & & & \\
& \quad \; -- N-back & \textbf{0.269 \(\pm\) 0.016} & \textbf{0.520 \(\pm\) 0.015} & \textbf{5.420 \(\pm\) 0.042} & \textbf{47.121 \(\pm\) 1.437} \\
& \quad \; -- MID & 0.198 \(\pm\) 0.007 & 0.447 \(\pm\) 0.007 & 5.687 \(\pm\) 0.086 & 51.673 \(\pm\) 1.445 \\
& \quad \; -- SST & 0.173 \(\pm\) 0.013 & 0.419 \(\pm\) 0.016 & 5.789 \(\pm\) 0.064 & 53.302 \(\pm\) 1.533 \\
\bottomrule
\end{tabular*}
\end{table*}

\begin{table*}[t]
\caption{Performance comparison of MAC-Net against conventional machine-learning and advanced deep-learning baselines. All models utilize N-back+MID tfMRI inputs and identical covariates under a common family-aware cross-validation framework.}
\label{tab:combined_benchmarks}
\centering
\small
\setlength{\tabcolsep}{4pt}
\begin{tabular*}{\textwidth}{@{\extracolsep{\fill}} l l c c c c}
\toprule
\textbf{Target} & \textbf{Model Configuration} & \textbf{\(R^2\) (\(\uparrow\))} & \textbf{Pearson \(r\) (\(\uparrow\))} & \textbf{MAE (\(\downarrow\))} & \textbf{MSE (\(\downarrow\))} \\
\midrule
\multirow{10}{*}{\textbf{\(G_f\)}} 
& \textbf{MAC-Net} & \textbf{0.174 \(\pm\) 0.009} & \textbf{0.420 \(\pm\) 0.009} & \textbf{6.855 \(\pm\) 0.046} & \textbf{74.748 \(\pm\) 1.373} \\
& \textit{Flattened Machine Learning} & & & & \\
& \quad \; -- Ridge & 0.105 \(\pm\) 0.020 & 0.357 \(\pm\) 0.018 & 7.132 \(\pm\) 0.100 & 80.939 \(\pm\) 3.230 \\
& \quad \; -- Lasso & 0.150 \(\pm\) 0.017 & 0.392 \(\pm\) 0.019 & 6.947 \(\pm\) 0.082 & 76.892 \(\pm\) 2.321 \\
& \quad \; -- ElasticNet & 0.151 \(\pm\) 0.015 & 0.395 \(\pm\) 0.018 & 6.933 \(\pm\) 0.105 & 76.814 \(\pm\) 2.577 \\
& \quad \; -- Random Forest & 0.119 \(\pm\) 0.018 & 0.348 \(\pm\) 0.030 & 7.070 \(\pm\) 0.163 & 79.718 \(\pm\) 3.194 \\
& \textit{Advanced Deep Learning} & & & & \\
& \quad \; -- Tabular Transformer & 0.108 \(\pm\) 0.014 & 0.331 \(\pm\) 0.021 & 7.121 \(\pm\) 0.127 & 80.662 \(\pm\) 2.275 \\
& \quad \; -- SpatialGATNet & 0.115 \(\pm\) 0.013 & 0.343 \(\pm\) 0.017 & 7.095 \(\pm\) 0.098 & 80.056 \(\pm\) 2.025 \\
& \quad \; -- Gated Tabular Net & 0.136 \(\pm\) 0.011 & 0.371 \(\pm\) 0.015 & 6.988 \(\pm\) 0.097 & 78.133 \(\pm\) 2.298 \\
\midrule
\multirow{10}{*}{\textbf{\(G_c\)}} 
& \textbf{MAC-Net} & \textbf{0.238 \(\pm\) 0.017} & \textbf{0.489 \(\pm\) 0.017} & \textbf{4.416 \(\pm\) 0.066} & \textbf{32.134 \(\pm\) 1.175} \\
& \textit{Flattened Machine Learning} & & & & \\
& \quad \; -- Ridge & 0.175 \(\pm\) 0.022 & 0.432 \(\pm\) 0.020 & 4.596 \(\pm\) 0.067 & 34.795 \(\pm\) 1.244 \\
& \quad \; -- Lasso & 0.205 \(\pm\) 0.007 & 0.454 \(\pm\) 0.008 & 4.500 \(\pm\) 0.045 & 33.506 \(\pm\) 0.770 \\
& \quad \; -- ElasticNet & 0.204 \(\pm\) 0.005 & 0.455 \(\pm\) 0.007 & 4.509 \(\pm\) 0.039 & 33.583 \(\pm\) 0.712 \\
& \quad \; -- Random Forest & 0.130 \(\pm\) 0.020 & 0.366 \(\pm\) 0.034 & 4.720 \(\pm\) 0.067 & 36.689 \(\pm\) 1.280 \\
& \textit{Advanced Deep Learning} & & & & \\
& \quad \; -- Tabular Transformer & 0.156 \(\pm\) 0.022 & 0.398 \(\pm\) 0.028 & 4.633 \(\pm\) 0.060 & 35.579 \(\pm\) 1.261 \\
& \quad \; -- SpatialGATNet & 0.175 \(\pm\) 0.022 & 0.422 \(\pm\) 0.024 & 4.598 \(\pm\) 0.074 & 34.802 \(\pm\) 1.415 \\
& \quad \; -- Gated Tabular Net & 0.178 \(\pm\) 0.014 & 0.422 \(\pm\) 0.016 & 4.593 \(\pm\) 0.044 & 34.676 \(\pm\) 1.037 \\
\midrule
\multirow{10}{*}{\textbf{\(G_t\)}} 
& \textbf{MAC-Net} & \textbf{0.277 \(\pm\) 0.008} & \textbf{0.527 \(\pm\) 0.008} & \textbf{5.394 \(\pm\) 0.036} & \textbf{46.580 \(\pm\) 0.973} \\
& \textit{Flattened Machine Learning} & & & & \\
& \quad \; -- Ridge & 0.214 \(\pm\) 0.021 & 0.473 \(\pm\) 0.017 & 5.620 \(\pm\) 0.079 & 50.642 \(\pm\) 1.611 \\
& \quad \; -- Lasso & 0.243 \(\pm\) 0.010 & 0.494 \(\pm\) 0.010 & 5.512 \(\pm\) 0.023 & 48.779 \(\pm\) 0.696 \\
& \quad \; -- ElasticNet & 0.240 \(\pm\) 0.013 & 0.492 \(\pm\) 0.012 & 5.528 \(\pm\) 0.024 & 48.994 \(\pm\) 0.717 \\
& \quad \; -- Random Forest & 0.177 \(\pm\) 0.014 & 0.425 \(\pm\) 0.021 & 5.744 \(\pm\) 0.078 & 53.081 \(\pm\) 1.676 \\
& \textit{Advanced Deep Learning} & & & & \\
& \quad \; -- Tabular Transformer & 0.177 \(\pm\) 0.010 & 0.423 \(\pm\) 0.012 & 5.749 \(\pm\) 0.080 & 53.026 \(\pm\) 1.542 \\
& \quad \; -- SpatialGATNet & 0.196 \(\pm\) 0.009 & 0.446 \(\pm\) 0.008 & 5.663 \(\pm\) 0.069 & 51.839 \(\pm\) 1.697 \\
& \quad \; -- Gated Tabular Net & 0.217 \(\pm\) 0.011 & 0.467 \(\pm\) 0.013 & 5.589 \(\pm\) 0.075 & 50.500 \(\pm\) 1.387 \\
\bottomrule
\end{tabular*}
\end{table*}

\section{Results and Discussion}


\subsection{Task-Specific Cognitive Modeling and Task Selection}
The covariate-only analysis (Table~\ref{tab:combined_baselines}) established that age, sex, scanner device, eTIV, and task-specific head motion accounted for 11.4\% to 17.8\% of the variance in uncorrected cognitive scores. This establishes a rigorous baseline: task-fMRI predictions must be evaluated against substantial demographic explanatory information rather than a null model.

Corresponding single-task MAC-Net analyses (Table~\ref{tab:combined_baselines}, bottom panels) revealed a consistent performance hierarchy across all three cognitive outcomes. N-back produced the highest mean \(R^2\) across \(G_f\), \(G_c\), and \(G_t\) (\(0.178 \pm 0.014\), \(0.223 \pm 0.022\), and \(0.269 \pm 0.016\), respectively). MID produced intermediate performance, whereas SST produced the lowest estimations across all domains. The stability of this ordering indicates that the predictive utility of the evaluated task representations was behaviorally task-dependent rather than determined solely by the available feature volume.

The strong performance of N-back is consistent with previous studies showing that cognitively demanding task states, particularly working-memory-related states, can be informative for individual differences in cognition \cite{greene2018task,chen2022shared,pat2023explainable,tetereva2022capturing}. Within the present representation, working-memory-related activation and contrast features therefore provided the most informative task-specific neural signal. This does not imply that cognitive ability is reducible to working memory, because the N-back response also reflects visual processing, sustained attention, response selection, motivation, and general task engagement.

MID retained non-trivial predictive information despite performing below N-back. Its inclusion in the multimodal configuration subsequently produced descriptive improvements for \(G_c\) and \(G_t\), suggesting that reward-related and outcome-monitoring task states may contain information complementary to working-memory-related activity. This observation is consistent with prior evidence that integrating task-fMRI information across paradigms can improve brain--cognition prediction \cite{tetereva2022capturing}. However, because the present task-selection procedure was informed by performance within the same cohort, the observed differences should be regarded as descriptive rather than selection-independent estimates.

SST was excluded from the primary multimodal configuration because it provided the weakest single-task performance across all three outcomes. The relatively weak performance may reflect the relevance of the selected inhibition contrasts to broad cognitive composites, the reliability of regional SST activation measures, the smaller feature space, or the sensitivity of the present architecture to this particular representation. Reliability limitations in ABCD task-fMRI measures have previously been reported \cite{kennedy2022reliability}. Thus, exclusion of SST is an empirical property of the present modeling pipeline rather than evidence that inhibitory-control circuitry is irrelevant to cognitive function.

The value of SST was further examined using a three-task sensitivity analysis. Adding SST to the primary N-back+MID configuration resulted in \(R^2=0.173\pm0.015\) for \(G_f\), \(0.233\pm0.015\) for \(G_c\), and \(0.273\pm0.014\) for \(G_t\), compared with \(0.174\pm0.009\), \(0.238\pm0.017\), and \(0.277\pm0.008\), respectively, for the N-back+MID configuration. The corresponding changes in mean \(R^2\) were approximately \(-0.001\), \(-0.005\), and \(-0.004\). These results provide no descriptive evidence that SST supplied incremental predictive information after N-back and MID had been incorporated. More broadly, the minor degradation in estimation performance suggests that injecting 658 additional regional features into the feed-forward architecture without providing substantial complementary variance exacerbates the curse of dimensionality. Rather than improving the model's descriptive capacity, the forced inclusion of these uninformative features complicates the optimization landscape. This highlights a critical design consideration for multi-task neural-signal systems: adding a task representation is beneficial only when it contributes unique predictive variance, rather than simply expanding the raw feature volume.

\begin{figure*}[t]
    \centering
    \includegraphics[width=\textwidth]{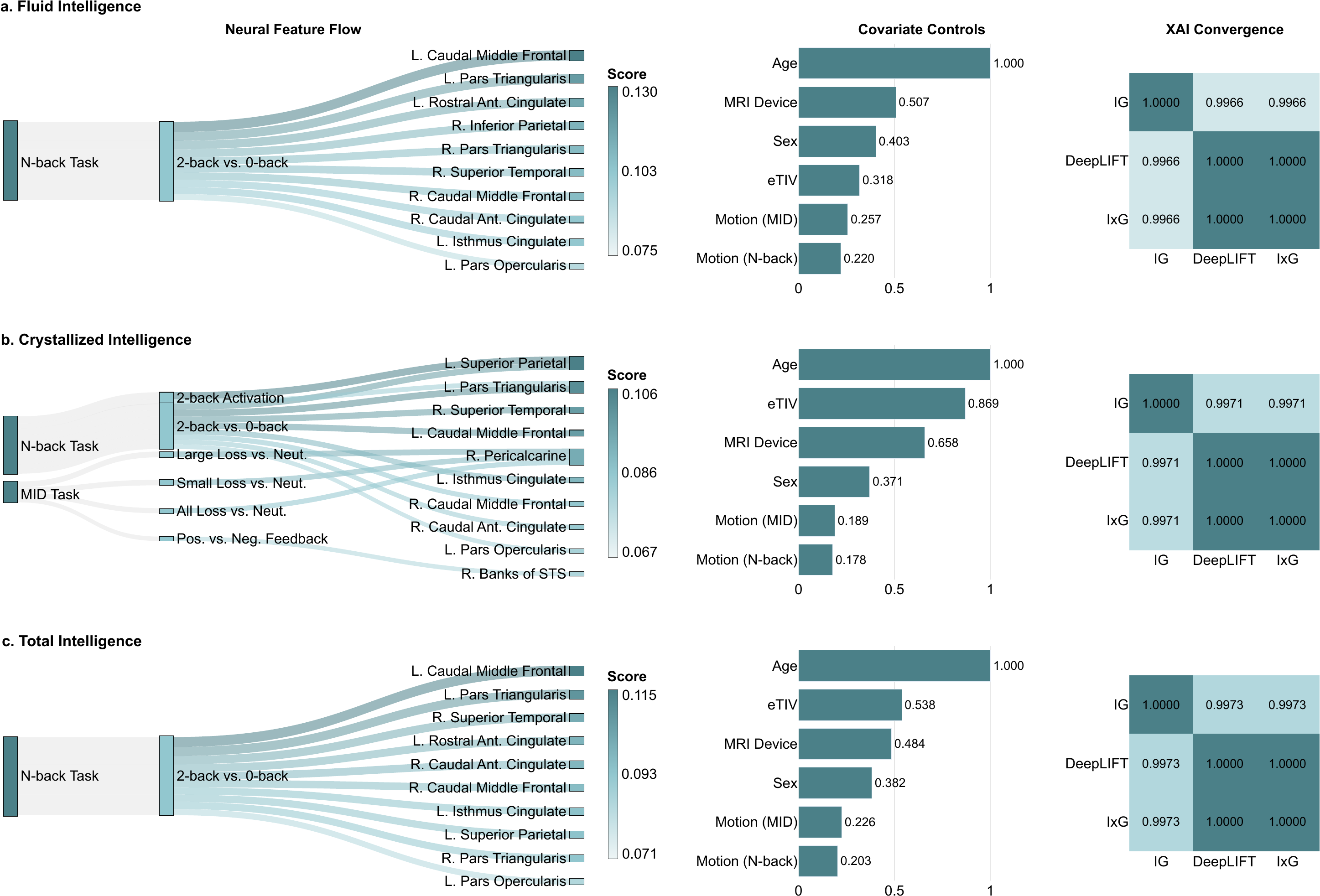}
    \caption{XAI attribution consensus for (a) fluid, (b) crystallized, and (c) total intelligence. Left panels depict the top ten task-condition-region features averaged across Integrated Gradients (IG), DeepLIFT, and Input\(\times\)Gradient (IxG). Center panels display normalized absolute covariate attributions (magnitude only). Right panels present pairwise Spearman rank correlations confirming multi-method agreement.}
    \label{fig:xai_consensus}
\end{figure*}

\subsection{Multi-Task Cognitive Estimation}
The primary N-back+MID MAC-Net configuration produced the strongest overall performance among all evaluated models (Table~\ref{tab:combined_benchmarks}). MAC-Net achieved mean \(R^2\) values of 0.174, 0.238, and 0.277 for \(G_f\), \(G_c\), and \(G_t\), respectively, consistently outperforming the strongest conventional regularized linear baselines (ElasticNet for \(G_f\), Lasso for \(G_c\) and \(G_t\)). These descriptive improvements represent absolute mean \(R^2\) gains of 0.023, 0.033, and 0.034 over conventional methods across the three respective outcomes.

Comparing single-task and multi-task configurations reveals phenotype-dependent integration benefits. For \(G_f\), N-back alone (\(R^2=0.178\pm0.014\)) performed similarly to the multi-task model (\(0.174\pm0.009\)). Conversely, the multi-task configuration exceeded N-back alone by 0.015 and 0.008 in mean \(R^2\) for \(G_c\) and \(G_t\), respectively, suggesting that broader composite cognition benefits more from complementary task states than fluid cognition under the present representation. Because participant-level paired uncertainty was not calculated, these improvements remain descriptive.

The brain-only ablation (detailed in Supplementary Table S2) further characterized the role of the two information domains. Removing all participant and acquisition covariates reduced MAC-Net's mean \(R^2\) to 0.110 for \(G_f\), 0.134 for \(G_c\), and 0.174 for \(G_t\). Crucially, this brain-only configuration still exceeded the Random Forest covariate-only baseline across all outcomes, confirming that tfMRI features provide independent predictive utility in the absence of demographic inputs. This pattern indicates complementary contributions from the task-fMRI and covariate feature spaces. Importantly, it does not constitute a formal decomposition of statistically independent neural and non-neural variance because imaging features may themselves contain information related to age, anatomy, motion, scanner characteristics, and other participant-level factors.

The observed performance pattern also supports the architectural motivation for late fusion. MAC-Net processes task-fMRI features through a dedicated neural pathway and introduces participant and acquisition variables only at the terminal prediction stage. This structural separation limits the direct influence of dominant covariates on early neural representation learning while retaining their predictive information for the final estimate. Such explicit control of information flow is potentially valuable for neural-signal modeling because participant and acquisition variables can otherwise dominate high-dimensional prediction problems.

Nevertheless, late fusion should not be interpreted as statistical deconfounding. Although measured covariates do not enter the early imaging pathway, the task-fMRI representation may still encode demographic, anatomical, motion-related, scanner-related, and other unmeasured sources of variation. The architecture therefore constrains information flow rather than guaranteeing a covariate-independent neural representation. This distinction is essential when interpreting the model as a neural engineering framework rather than as a causal or deconfounding procedure.

\subsection{Architectural Benchmarking}
MAC-Net achieved the highest mean performance among all evaluated deep-learning architectures (Table~\ref{tab:combined_benchmarks}). The Gated Tabular Network served as the strongest alternative (mean \(R^2\) of 0.136, 0.178, and 0.217 for \(G_f\), \(G_c\), and \(G_t\)), outperforming both the SpatialGATNet and the Tabular Transformer. Ultimately, MAC-Net's static feed-forward pathways exceeded the strongest alternative deep model by 0.038, 0.060, and 0.060 in mean \(R^2\) across the three respective outcomes.

The consistent performance advantage of MAC-Net indicates that its feed-forward representation and late-fusion design were well matched to the atlas-averaged regional task-fMRI features used in this study. This result is noteworthy because conventional regularized linear models remain strong baselines for brain--behavior prediction, and previous work has shown that increasingly complex deep-learning architectures do not necessarily improve individual-difference modeling \cite{schulz2020different,pat2023explainable,popov2024simple}. The present findings therefore favor an architecture that is appropriately matched to the structure and dimensionality of the input representation rather than maximal architectural complexity.

The lower performance of SpatialGATNet and the Tabular Transformer does not establish that graph or self-attention mechanisms are intrinsically unsuitable for task-fMRI. Their efficacy depends heavily on graph construction (e.g., our fixed \(K=10\) MNI prior), tokenization, positional encoding, and hyperparameter optimization. Comprehensive benchmarking across alternative topological priors and matched parameter budgets is required before drawing broader conclusions regarding these model classes.

\begin{figure*}[t]
    \centering
    \includegraphics[width=\textwidth]{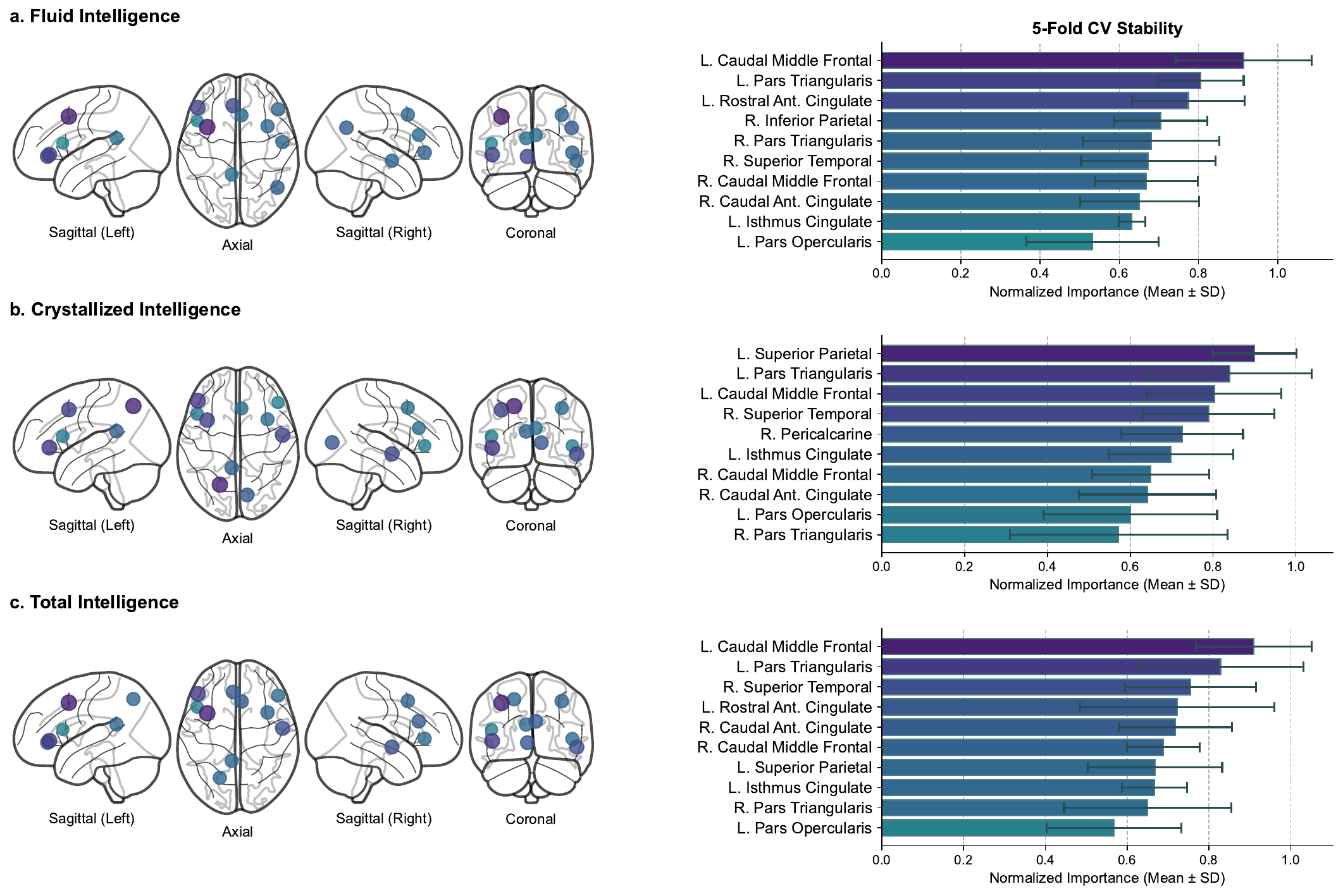}
    \caption{Spatial distribution and fold-wise stability of the top ten model-attributed regions for (a) fluid, (b) crystallized, and (c) total intelligence. Left panels show MNI projections of regional hubs, scaled by explanatory importance. Right panels detail normalized mean Integrated Gradients (IG) scores, with error bars denoting standard deviation across the five cross-validation folds.}
    \label{fig:xai_stability}
\end{figure*}

\subsection{Model Attribution and Neural Feature Stability}

The attribution analysis provided a complementary assessment of how MAC-Net formed its predictions (Fig. ~\ref{fig:xai_consensus}). Integrated Gradients, DeepLIFT, and Input\(\times\)Gradient produced exceptionally high agreement in global feature rankings. Pairwise Spearman correlations exceeded \(\rho=0.996\) across all explainer combinations and cognitive targets (\(G_f\), \(G_c\), \(G_t\)). This convergence indicates that the global ranking of model-attributed features was highly robust to the choice of gradient-based attribution method under the selected common zero-reference baseline.

The highest-ranked covariate attribution was age across all three cognitive outcomes. Scanner device, eTIV, sex, and task-specific motion also contributed substantial attribution magnitude, although their relative ordering differed by outcome. Because the cognitive targets were analyzed as uncorrected NIH Toolbox scores in a developing population, the dominant contribution of age is expected and highlights the importance of developmental variation in the prediction problem. Importantly, these normalized absolute attribution values quantify model sensitivity rather than the direction or causal effect of a variable.

Within the neural feature space, the highest-ranked features for \(G_f\) and \(G_t\) were dominated by the N-back 2-back versus 0-back contrast. Prominent regions included the left caudal middle frontal gyrus, left pars triangularis, rostral and caudal anterior cingulate cortices, and right inferior and superior parietal regions. This spatial pattern broadly overlaps with distributed frontal--parietal systems implicated in working memory and higher-order cognitive control \cite{jung2007pfit,deary2010neuroscience,bassett2017network}. The convergence between the predictive analysis and established neurocognitive models provides a biologically plausible interpretation of the features used by MAC-Net.

For \(G_c\), the attribution profile integrated N-back features with MID loss-anticipation and feedback signals in pericalcarine, cingulate, and frontal regions. Because MID does not directly measure crystallized knowledge, this likely reflects the model capturing shared variance in attention, motivation, or generalized task engagement predictive of \(G_c\) within this cohort, rather than a specific reward-based mechanism.

The cross-validation stability analysis further showed that the leading attributed regions retained non-zero mean importance across all five folds (Fig. ~\ref{fig:xai_stability}). However, the observed fold-to-fold variation indicates that exact feature rankings remain sensitive to sampling. The broad spatial distribution of leading features is therefore more reproducible than their precise hierarchical ordering.

Although the strong agreement among attribution methods is encouraging, it does not establish biological validity. Integrated Gradients, DeepLIFT, and Input\(\times\)Gradient are all gradient-based approaches applied to the same model and reference convention, and highly correlated features can exchange attribution magnitude without materially changing predictions \cite{kohoutova2020interpreting,adebayo2018sanity}. Recent evidence further indicates that alternative feature-selection procedures can produce substantially different neurobiological interpretations despite similar predictive performance \cite{adkinson2026feature}. Accordingly, the features identified here should be regarded as model-supported neural correlates rather than definitive biomarkers or causal mechanisms.

\subsection{Limitations and Translational Implications}
Several limitations guide the interpretation of these findings. First, the analysis utilized cross-sectional baseline data from a single developmental cohort; future work must leverage the longitudinal structure of the ABCD Study to model within-person cognitive trajectories. Second, because task selection was informed by predictive performance within the same cohort and lacked formal paired statistical testing, the reported performance differences remain descriptive. Nested cross-validation and participant-level paired inference (e.g., family-clustered bootstrapping) are required to yield fully selection-independent generalization estimates. Third, strict complete-case filtering ensured data integrity but may introduce sociodemographic selection biases, necessitating future evaluations under alternative missing-data strategies. Fourth, our computationally efficient reliance on atlas-averaged beta estimates inherently discards fine-grained spatial variation and temporal dynamics. Given the documented reliability limits of tfMRI \cite{gell2024measurement,kennedy2022reliability}, future research must empirically justify whether more complex voxelwise or dynamic representations yield reproducible performance gains. Finally, MAC-Net was developed within a healthy cohort, establishing a methodological foundation rather than immediate clinical utility. Translational deployment for objective cognitive evaluation \cite{wen2022taskstate,kuo2023optimized} will require independent multi-site validation and rigorous calibration within neurologically impaired and rehabilitation populations.

\section{Conclusion}
This study presented MAC-Net, a covariate-aware multi-task deep-learning framework that isolates task-fMRI representation learning from participant and acquisition variables via late fusion. Evaluated on 6,500 ABCD Study participants under strict family-aware cross-validation, MAC-Net consistently outperformed conventional machine learning and advanced deep-learning baselines. Structured late-fusion effectively extracted incremental neural signals beyond demographic confounders, with brain-only ablations confirming predictive utility even without covariates. N-back provided the strongest single-task representation, and integrating MID established a parsimonious, high-performing multi-task configuration.

Furthermore, multi-method explainability analyses demonstrated exceptional concordance, localizing critical predictive features to N-back-related frontal, parietal, and cingulate regions. While these attributions represent associative correlates rather than causal biomarkers, they confirm the model's reliance on biologically plausible networks. 

Overall, MAC-Net establishes a leakage-resistant, interpretable computational framework for estimating cognitive function from task-evoked neural signals. Reliably integrating heterogeneous data streams provides a methodological foundation for future neural engineering applications. Prospective multi-site validation and longitudinal testing in neurologically impaired populations will be the next critical steps toward deploying such objective cognitive characterization tools in clinical and neurorehabilitation settings.

\section*{Acknowledgment}
This research was supported by the Science and Technology Fellowship Trust, Government of the People's Republic of Bangladesh, and the Commonwealth through an Australian Government Research Training Program Scholarship. Data used in the preparation of this article were obtained from the Adolescent Brain Cognitive Development (ABCD) Study, held in the NIH Brain Development Cohorts Data Sharing Platform. In addition, this work was supported by resources provided by The University of Queensland Research Computing Center’s Bunya supercomputer.

\printbibliography


\clearpage
\onecolumn
\appendices
\setcounter{page}{1}
\setcounter{section}{0}
\setcounter{figure}{0}
\setcounter{table}{0}
\renewcommand{\thesection}{S\arabic{section}}
\renewcommand{\thefigure}{S\arabic{figure}}
\renewcommand{\thetable}{S\arabic{table}}

\begin{center}
    \Large\textbf{Supplementary Material: A Multi-Task Deep Learning Framework for Modeling Cognitive Function From Task-Based fMRI}
\end{center}

\vspace{0.5cm}

\noindent\textbf{Section S1: Anatomical Regions of Interest}
\vspace{0.2cm}

\noindent The tfMRI functional representations used in this study were extracted from 94 predefined anatomical regions. 

\vspace{0.2cm}
\noindent\textbf{Cortical Regions (n = 68, 34 per hemisphere):} Banks of the Superior Temporal Sulcus, Caudal Anterior Cingulate, Caudal Middle Frontal, Cuneus, Entorhinal, Fusiform, Inferior Parietal Lobe, Inferior Temporal, Isthmus Cingulate, Lateral Occipital, Lateral Orbitofrontal, Lingual, Medial Orbitofrontal, Middle Temporal, Parahippocampal, Paracentral, Pars Opercularis, Pars Orbitalis, Pars Triangularis, Pericalcarine, Postcentral, Posterior Cingulate, Precentral, Precuneus, Rostral Anterior Cingulate, Rostral Middle Frontal, Superior Frontal, Superior Parietal, Superior Temporal, Supramarginal, Frontal Pole, Temporal Pole, Transverse Temporal, and the Insula.
 
\vspace{0.2cm}
\noindent\textbf{ASEG Anatomical Segments (n = 26, 13 per hemisphere):} Cerebral White Matter, Lateral Ventricle, Inferior Lateral Ventricle, Cerebellar White Matter, Cerebellar Cortex, Thalamus Proper, Caudate, Putamen, Pallidum, Hippocampus, Amygdala, Accumbens Area, and the Ventral Diencephalon.

\vspace{0.5cm}

\begin{table}[h!]
\caption{Summary of task-fMRI conditions and functional contrasts included in the analysis. Beta estimates were extracted for each condition or contrast across 94 anatomical regions (comprising 68 Desikan--Killiany cortical parcels and 26 bilateral ASEG anatomical segments). This extraction yielded 846 features for N-back, 940 for the Monetary Incentive Delay (MID) task, and 658 for the Stop Signal Task (SST).}
\label{tab:tfmri_features_supp}
\centering
\footnotesize
\begin{tabularx}{\textwidth}{@{} p{1.5cm} p{6.5cm} X c @{}}
\toprule
\textbf{Task} & \textbf{Condition / Contrast} & \textbf{Psychological Process} & \textbf{Total Features} \\
\midrule
\multirow{9}{*}{\textbf{N-Back}} 
 & 0-back (Baseline) & Baseline attention and visual processing. & \multirow{9}{*}{846} \\
 & 2-back (Load) & High working memory load. & \\
 & 2-back vs. 0-back & Pure working memory updating signature. & \\
 & Places & Visual processing of place stimuli. & \\
 & Emotional Faces & Visual processing of emotional stimuli. & \\
 & Face vs. Place & Category-specific visual processing. & \\
 & Emotional vs. Neutral & General emotional reactivity. & \\
 & Positive vs. Neutral & Positive emotional reactivity. & \\
 & Negative vs. Neutral & Negative emotional regulation. & \\
\midrule
\multirow{10}{*}{\textbf{MID}} 
 & Anticipation Reward vs. Neutral & General reward anticipation. & \multirow{10}{*}{940} \\
 & Anticipation Large Reward vs. Neutral & High reward anticipation. & \\
 & Anticipation Small Reward vs. Neutral & Low reward anticipation. & \\
 & Anticipation Loss vs. Neutral & General punishment anticipation. & \\
 & Anticipation Large Loss vs. Neutral & High loss anticipation. & \\
 & Anticipation Small Loss vs. Neutral & Low loss anticipation. & \\
 & Anticipation Large vs. Small Reward & Reward sensitivity scaling. & \\
 & Anticipation Large vs. Small Loss & Loss sensitivity scaling. & \\
 & Reward Positive vs. Negative Feedback & Reward outcome receipt. & \\
 & Loss Positive vs. Negative Feedback & Loss avoidance receipt. & \\
\midrule
\multirow{7}{*}{\textbf{SST}} 
 & Correct Go vs. Fixation & Basic motor response execution. & \multirow{7}{*}{658} \\
 & Correct Stop vs. Correct Go & Successful motor inhibition. & \\
 & Correct Stop vs. Incorrect Stop & Successful vs. failed inhibition. & \\
 & Incorrect Go vs. Correct Go & Motor execution error. & \\
 & Incorrect Go vs. Incorrect Stop & Execution error vs. inhibition failure. & \\
 & Incorrect Stop vs. Correct Go & Failure to inhibit motor response. & \\
 & Any Stop vs. Correct Go & General stop-signal processing. & \\
\bottomrule
\end{tabularx}

\parbox{\textwidth}{\footnotesize \textit{Note:} N-back = N-back working memory task; MID = Monetary Incentive Delay task; SST = Stop Signal Task. DK = Desikan-Killiany cortical atlas (68 nodes: 34 per hemisphere). ASEG = Automatic Subcortical Segmentation (26 nodes: 13 per hemisphere). Total features per task = (Number of conditions/contrasts) \(\times\) 94, where 94 = 68 (DK cortical) + 26 (ASEG subcortical).}
\end{table}

\clearpage

\begin{table}[t!]
\caption{Brain-only ablation performance using N-back+MID tfMRI inputs without participant or acquisition covariates. Values reflect mean \(\pm\) standard deviation across five held-out test folds.}
\label{tab:brain_only_supp}
\centering
\small
\begin{tabularx}{\textwidth}{@{\extracolsep{\fill}} l l c c c c}
\toprule
\textbf{Target} & \textbf{Model} & \textbf{\(R^2\) (\(\uparrow\))} & \textbf{Pearson \(r\) (\(\uparrow\))} & \textbf{MAE (\(\downarrow\))} & \textbf{MSE (\(\downarrow\))} \\
\midrule
\multirow{5}{*}{\textbf{\(G_f\)}} 
& Ridge & 0.045 \(\pm\) 0.022 & 0.284 \(\pm\) 0.022 & 7.382 \(\pm\) 0.133 & 86.447 \(\pm\) 3.617 \\
& Lasso & 0.087 \(\pm\) 0.017 & 0.306 \(\pm\) 0.020 & 7.200 \(\pm\) 0.114 & 82.599 \(\pm\) 2.753 \\
& ElasticNet & 0.090 \(\pm\) 0.019 & 0.310 \(\pm\) 0.024 & 7.183 \(\pm\) 0.120 & 82.366 \(\pm\) 3.027 \\
& Random Forest & 0.084 \(\pm\) 0.008 & 0.293 \(\pm\) 0.015 & 7.217 \(\pm\) 0.137 & 82.904 \(\pm\) 2.529 \\
& \textbf{MAC-Net} & \textbf{0.110 \(\pm\) 0.010} & \textbf{0.341 \(\pm\) 0.011} & \textbf{7.107 \(\pm\) 0.125} & \textbf{80.532 \(\pm\) 2.436} \\
\midrule
\multirow{5}{*}{\textbf{\(G_c\)}} 
& Ridge & 0.093 \(\pm\) 0.032 & 0.345 \(\pm\) 0.029 & 4.819 \(\pm\) 0.094 & 38.239 \(\pm\) 1.439 \\
& Lasso & 0.121 \(\pm\) 0.016 & 0.354 \(\pm\) 0.018 & 4.733 \(\pm\) 0.050 & 37.045 \(\pm\) 0.725 \\
& ElasticNet & 0.119 \(\pm\) 0.013 & 0.350 \(\pm\) 0.021 & 4.733 \(\pm\) 0.039 & 37.126 \(\pm\) 0.689 \\
& Random Forest & 0.075 \(\pm\) 0.018 & 0.279 \(\pm\) 0.037 & 4.880 \(\pm\) 0.075 & 38.994 \(\pm\) 1.279 \\
& \textbf{MAC-Net} & \textbf{0.134 \(\pm\) 0.015} & \textbf{0.379 \(\pm\) 0.018} & \textbf{4.742 \(\pm\) 0.053} & \textbf{36.514 \(\pm\) 0.995} \\
\midrule
\multirow{5}{*}{\textbf{\(G_t\)}} 
& Ridge & 0.122 \(\pm\) 0.030 & 0.380 \(\pm\) 0.026 & 5.925 \(\pm\) 0.095 & 56.570 \(\pm\) 2.174 \\
& Lasso & 0.150 \(\pm\) 0.017 & 0.392 \(\pm\) 0.018 & 5.826 \(\pm\) 0.049 & 54.774 \(\pm\) 0.941 \\
& ElasticNet & 0.145 \(\pm\) 0.019 & 0.388 \(\pm\) 0.021 & 5.833 \(\pm\) 0.063 & 55.096 \(\pm\) 1.225 \\
& Random Forest & 0.118 \(\pm\) 0.013 & 0.347 \(\pm\) 0.023 & 5.940 \(\pm\) 0.082 & 56.879 \(\pm\) 1.677 \\
& \textbf{MAC-Net} & \textbf{0.174 \(\pm\) 0.013} & \textbf{0.424 \(\pm\) 0.012} & \textbf{5.758 \(\pm\) 0.064} & \textbf{53.224 \(\pm\) 0.974} \\
\bottomrule
\end{tabularx}
\end{table}

\end{document}